\documentclass[10pt,twocolumn,letterpaper]{article}

\usepackage[pagenumbers]{cvpr} 
\usepackage{multirow} 
\usepackage{graphicx}
\usepackage{bbding}

\definecolor{cvprblue}{rgb}{0.21,0.49,0.74}
\usepackage[pagebackref,breaklinks,colorlinks,allcolors=cvprblue]{hyperref}

\def\paperID{500} 
\def\confName{CVPR}
\def\confYear{2025}

\title{MonoDETRNext: Next-Generation Accurate and Efficient Monocular 3D Object Detector}

\author{
	Liao Pan,
	Yang Feng*,
	Wu Di,
	Zhao Wenhui,
	Yu Jinwen
	}

\begin{document}
\maketitle

\begin{abstract}
Monocular 3D object detection has vast application potential across various fields. DETR-type models have shown remarkable performance in different areas, but there is still considerable room for improvement in monocular 3D detection, especially with the existing DETR-based method, MonoDETR. After addressing the query initialization issues in MonoDETR, we explored several performance enhancement strategies, such as incorporating a more efficient encoder and utilizing a more powerful depth estimator. Ultimately, we proposed MonoDETRNext, a model that comes in two variants based on the choice of depth estimator: MonoDETRNext-E, which prioritizes speed, and MonoDETRNext-A, which focuses on accuracy. We posit that MonoDETRNext establishes a new benchmark in monocular 3D object detection and opens avenues for future research. We conducted an exhaustive evaluation demonstrating the model’s superior performance against existing solutions. Notably, MonoDETRNext-A demonstrated a 3.52$\%$ improvement in the $AP_{3D}$ metric on the KITTI test benchmark over MonoDETR, while MonoDETRNext-E showed a 2.35$\%$ increase. Additionally, the computational efficiency of MonoDETRNext-E slightly exceeds that of its predecessor.
\end{abstract}    
\footnotetext{Liao Pan, Yang Feng, Wu Di,
	Yu Jinwen,
	Zhao Wenhui are with Key Laboratory of Information Fusion, School of Automation, Northwestern Polytechnical University,Xi'an,China.	(e-mile: liaopan@mail.nwpu.edu.cn,yangfeng@nwpu.edu.cn,wu\_di821@mail.
	nwpu.edu.cn,
	yujinwen@mail.nwpu.edu.cn, zwh2024202513@mail.nwpu.edu.cn}
\section{Introduction}
3D object detection boasts a vast range of applications, from autonomous driving and robotic navigation to smart surveillance and virtual reality. These applications hinge critically on the precise recognition and localization of objects within 3D space. However, many 3D object detection methods often rely on expensive equipment such as LiDAR \cite{wu2023virtual,dong2023pep,li2023logonet,wang2023object,hu2023ea,liang2022bevfusion}, significantly limiting their practical deployment. While several purely visual strategies have achieved notable success, these generally involve multiple cameras \cite{li2022bevformer,yang2023bevformer,pan2024clip} However, these approaches are often multi-camera based, and they can be limited in certain situations, such as when a LiDAR or one or more cameras suddenly malfunction. Additionally, they generally require substantial computational power for support, which further reinforces their limitations. Monocular 3D detection methods, on the other hand, do not have these constraints. They can achieve satisfactory performance with just a single camera, making this class of methods highly valuable for research.


DETR-based models have demonstrated remarkable performance in both 2D detection \cite{carion2020end,zhu2020deformable,zhang2022dino,chen2023group,zong2023detrs} and multi-view \cite{li2022bevformer,yang2023bevformer,zhang2024sparselif,wang2022detr3d}, multi-source 3D detection \cite{chen2023futr3d,liang2022bevfusion,brazil2023omni3d} . MonoDETR \cite{zhang2023monodetr} was the first end-to-end monocular 3D object detection model in the DETR category. Upon its introduction, its performance surpassed all previous monocular 3D detection approaches. However, MonoDETR still exhibited certain limitations, with two notable issues: during query initialization, both the encoding and reference points lacked physical significance, and the depth estimation network was overly simplistic.

To address these issues, we first conducted an analysis of query generation for monocular 3D tasks and proposed a novel monocular 3D query generation method. This approach derives reference points and query encodings more effectively from depth and image features, significantly enhancing performance. Concurrently, we implemented a more efficient encoder to improve the model's feature extraction capabilities while marginally increasing inference speed. These series of improvements led to the development of MonoDETRNext-E.

Subsequently, after experimenting with multiple established depth estimation networks, we identified a network that effectively enhanced MonoDETR's detection performance. Based on this network and the aforementioned improvements, we introduced MonoDETRNext-A, a model with robust detection performance but slightly lower inference speed.

%

Through this work, we aim to provide a significant starting point for the future development of monocular 3D vision detection models, laying a solid foundation for subsequent research and applications.

The salient innovations encapsulated within this manuscript may be delineated as follows:
\begin{itemize}
	\item [1)]
	Proposing  two novel monocular 3D object detection models, namely MonoDETRNext-E and MonoDETRNext-A, the former adeptly balancing speed and precision, while the latter accentuates precision-centric objectives.
	\item [2)]
Addressing the issue of nonsensical query initialization in MonoDETR, we introduced a bespoke object query generation strategy tailored for monocular 3D object detection, thereby refining model performance.

	\item [3)]
Evaluating numerous advanced networks, we identified two that were particularly effective. These findings facilitated the development of a new hybrid encoder and a streamlined, accurate depth estimation module for MonoDETR, significantly enhancing the model's efficiency and detection precision.
\end{itemize}

\section{Related Work}
\label{relate_work}

Current 3D object detection methods can generally be classified into two categories: camera-based methods and fusion methods integrating LiDAR and other sensors.

Camera-based methods can be further divided into monocular (single-view) and multi-view methods based on the number of input viewpoints. Monocular detectors utilize only forward-facing images as input, addressing more complex tasks with limited 2D information. Multi-view detectors simultaneously encode images of the surrounding scene, leveraging relationships between viewpoints to understand 3D space. On the other hand, fusion methods based on LiDAR and other sensor integration rely on inputs from devices such as depth cameras and LiDAR, which provide a fusion of various sensor data types, including images and point clouds. Consequently, they can acquire richer and more comprehensive depth information. Despite their higher cost, these methods typically exhibit greater robustness and accuracy in complex environments, as they can exploit the advantages of multiple sensors and integrate information from different data sources.

\subsection{MonoDETR and other monocular 3D detection}

MonoDETR \cite{zhang2023monodetr} is a state-of-the-art method that leverages rendered transmittance to predict depth maps from a single RGB image. By capturing subtle cues in the input image, MonoDETR achieves improved accuracy and robustness to varying lighting conditions compared to traditional monocular depth estimation methods. 

Several other monocular 3D reconstruction method have been proposed in recent years. For example, MonoDTR\cite{huang2022monodtr} is a deep learning model that predicts depth maps from single RGB images using a transformer-based architecture. While MonoDTR achieves high accuracy, it requires additional LiDAR data for training assistance. Meanwhile, CaDDN\cite{reading2021categorical} and Monorun\cite{chen2021monorun} not only require LiDAR data during training but also during inference. Autoshape\cite{liu2021autoshape}  integrates CAD data into the model to augment the restricted 3D cues. MonoDETR requires minimal 2D-3D geometric priors and does not necessitate additional annotations. Our MonoDETRNext inherits this characteristic.

Alternative methodologies, as demonstrated by MonoDLE \cite{ma2021delving}, PGD \cite{wang2022probabilistic}, and PackNet \cite{mallya2018packnet}, integrate multi-scale feature fusion and attention mechanisms for depth map estimation and error analysis, yielding improved performance. Despite their high accuracy, these approaches incur substantial computational costs and demand significant memory resources. Conversely, MonoDETR is characterized by its lightweight and efficient nature. Furthermore, MonoDETRNext-E surpasses it in speed and efficacy, while MonoDETRNext-A demonstrates markedly superior performance.

\subsection{Multi-view 3D object detection}

\begin{figure}
	\centering
	\label{different_3d_model}
	\includegraphics[width=1\columnwidth]{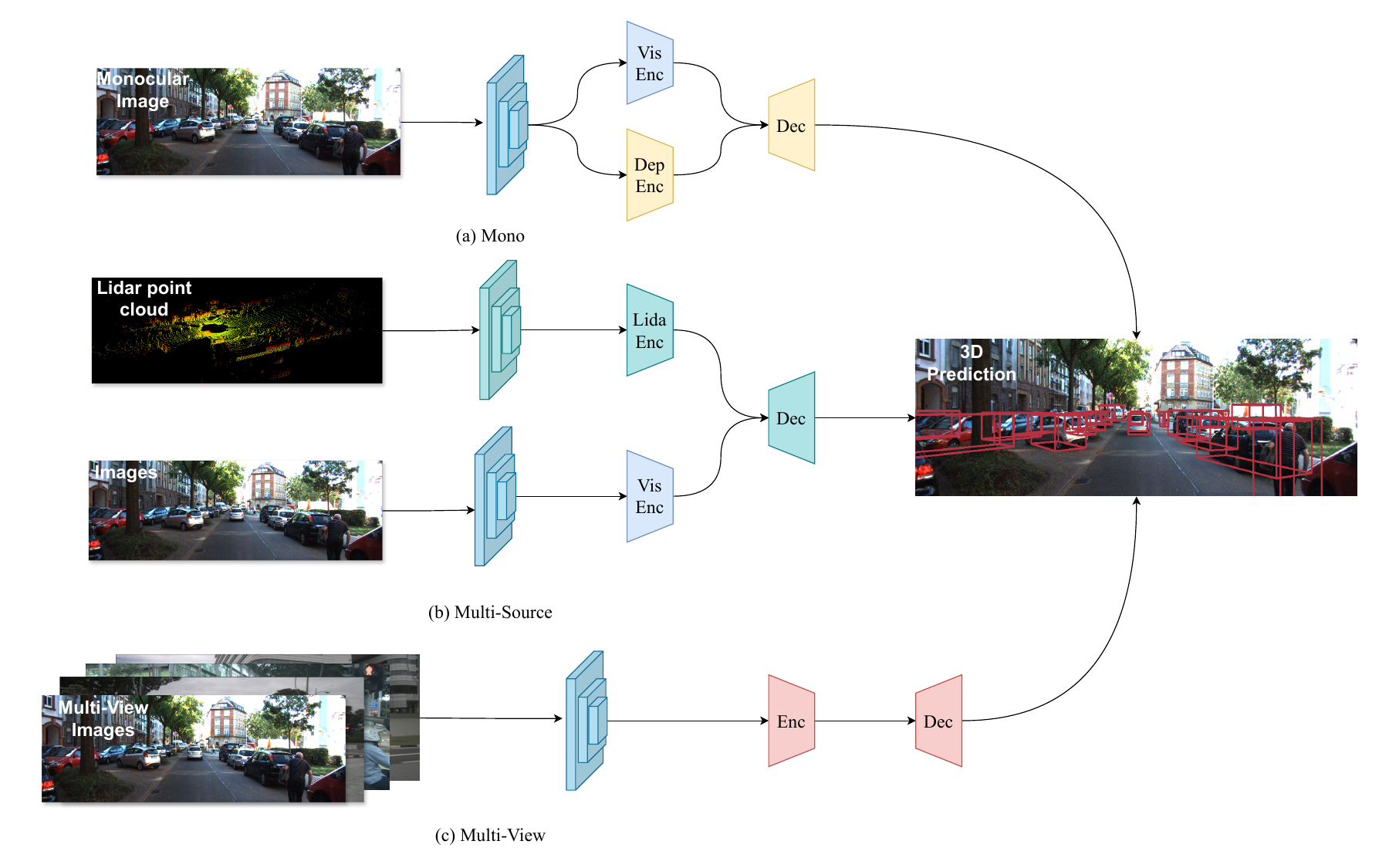}
	\caption{
		Comparison of DETR-type 3D detection models, with different colors representing distinct functional modules.}
\end{figure}

To jointly extract features from surrounding views, DETR3D \cite{wang2022detr3d} initially employs a set of 3D object queries, which are then back-projected onto multi-view images to aggregate features. The PETR series \cite{liu2022petr,liu2023petrv2,wang2023exploring} further introduces the generation of 3D positional features, avoiding unstable projections, and explores the advantages of temporal information from the preceding frame.

Alternatively, BEVFormer \cite{li2022bevformer} and its improvements \cite{yang2023bevformer,pan2024clip} generate BEV (Bird's Eye View) features using learnable BEV queries and introduce a spatiotemporal BEV transformer for visual feature aggregation. Subsequent research has also investigated cross-modal distillation \cite{huang2022tig,zhao2024simdistill} and masked image modeling \cite{yang2022towards,chen2023pimae} to enhance performance.

\subsection{LiDAR and multi-source information fusion 3D object detection}

Methods such as DeepFusion \cite{li2022deepfusion} and PointPainting \cite{vora2020pointpainting} represent notable advancements in the integration of LiDAR point cloud data with camera imagery to facilitate precise object detection within three-dimensional spatial environments. This fusion strategy optimally exploits the synergies inherent in disparate sensor modalities, amalgamating spatial depth cues with color texture information, thereby fortifying the resilience and accuracy of detection outcomes.
The integration of principles from BEVFormer\cite{li2022bevformer} into fusion paradigms, exemplified by BevFusion \cite{liang2022bevfusion}, has spurred further refinements culminating in enhanced precision, as evidenced by exemplar models such as those delineated in MV2D \cite{wang2023object} and Futr3d \cite{chen2023futr3d}. Recent endeavors, typified by mmFusion \cite{ahmad2023mmfusion}, have expanded the purview of fusion methodologies by integrating data from multiple sensors, including cameras, LiDAR, and radar, resulting in notable performance strides.

Concurrently, the domain has witnessed the emergence of large-scale architectures, exemplified by OMNI3D \cite{brazil2023omni3d} and GLEE \cite{wu2023general}, which have showcased remarkable efficacy in 3D object detection tasks. 
\section{Method}
\label{method}

\subsection{Model Overview}

\begin{figure*}
	\centering
	\includegraphics[width=0.8\textwidth]{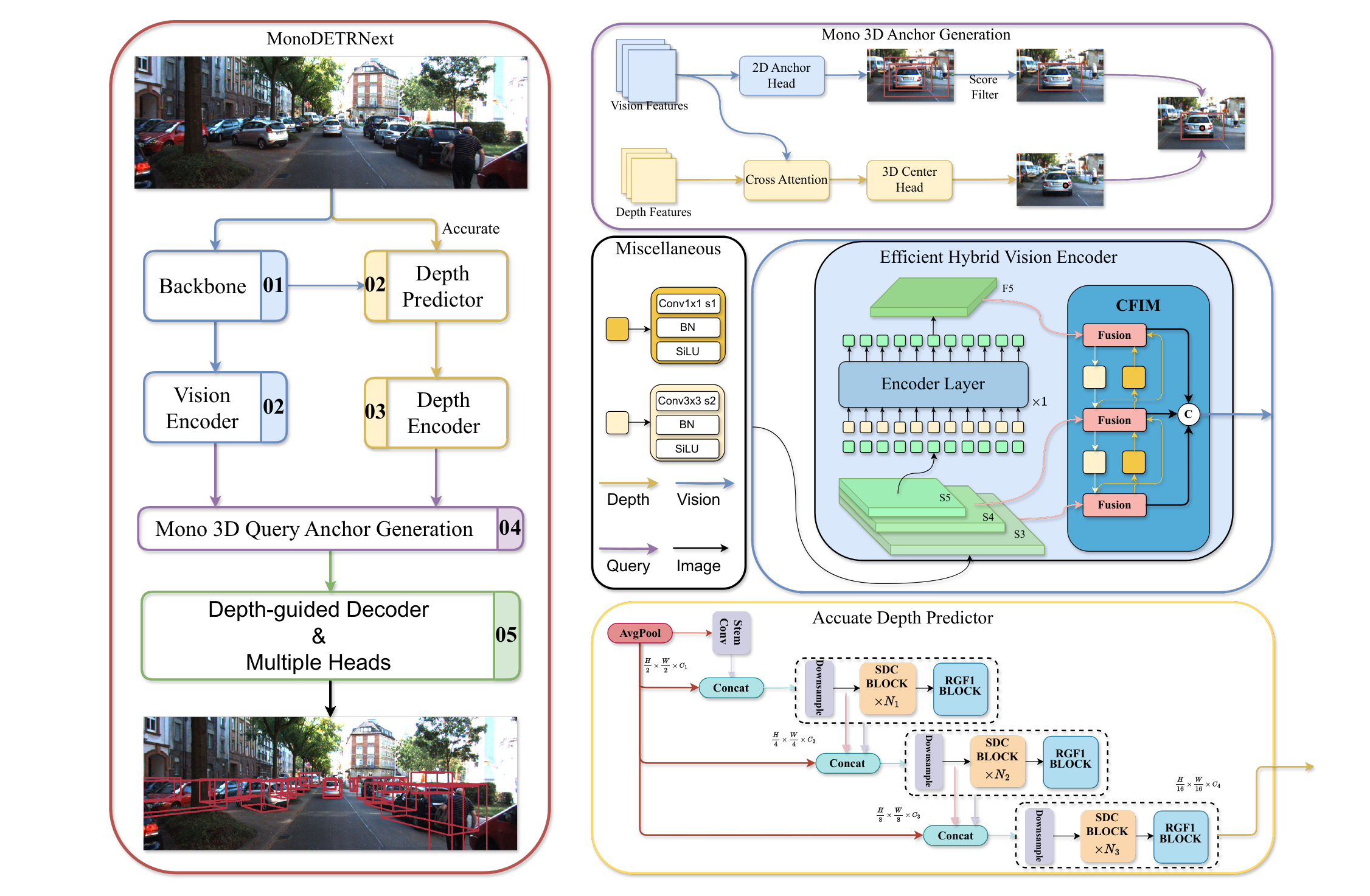}
	\caption{
		\textbf{The schematic depiction of MonoDETRNext.} The distinction between MonoDETRNext-A and MonoDETRNext-E primarily resides in their respective depth prediction mechanisms. The provided illustration delineates the intricate depth prediction scheme adopted by MonoDETRNext-A, whereas the depth predictor employed in MonoDETRNext-E remains congruent with that of MonoDETR.}
	\label{framework}
\end{figure*}

The overall frameworks of MonoDETRNext-A and MonoDETRNext-E are illustrated in Figure \ref{framework}. A key focus of this paper is the meticulous generation of object queries, along with the associated loss, which will be expounded in Section \ref{query}. The main difference between these two models lies in the disparity of the depth predictors, which will be elaborated in Section \ref{depth_predictor}. Additionally, the design principles of the efficient encoder and the extraction of visual and depth features will be explicated in Section \ref{encoder}.

\subsection{MonoDETRNext-E}

\subsubsection{Depth-Guided Query Anchors}

\label{query}
\begin{figure}
	\centering
	\includegraphics[width=1\columnwidth]{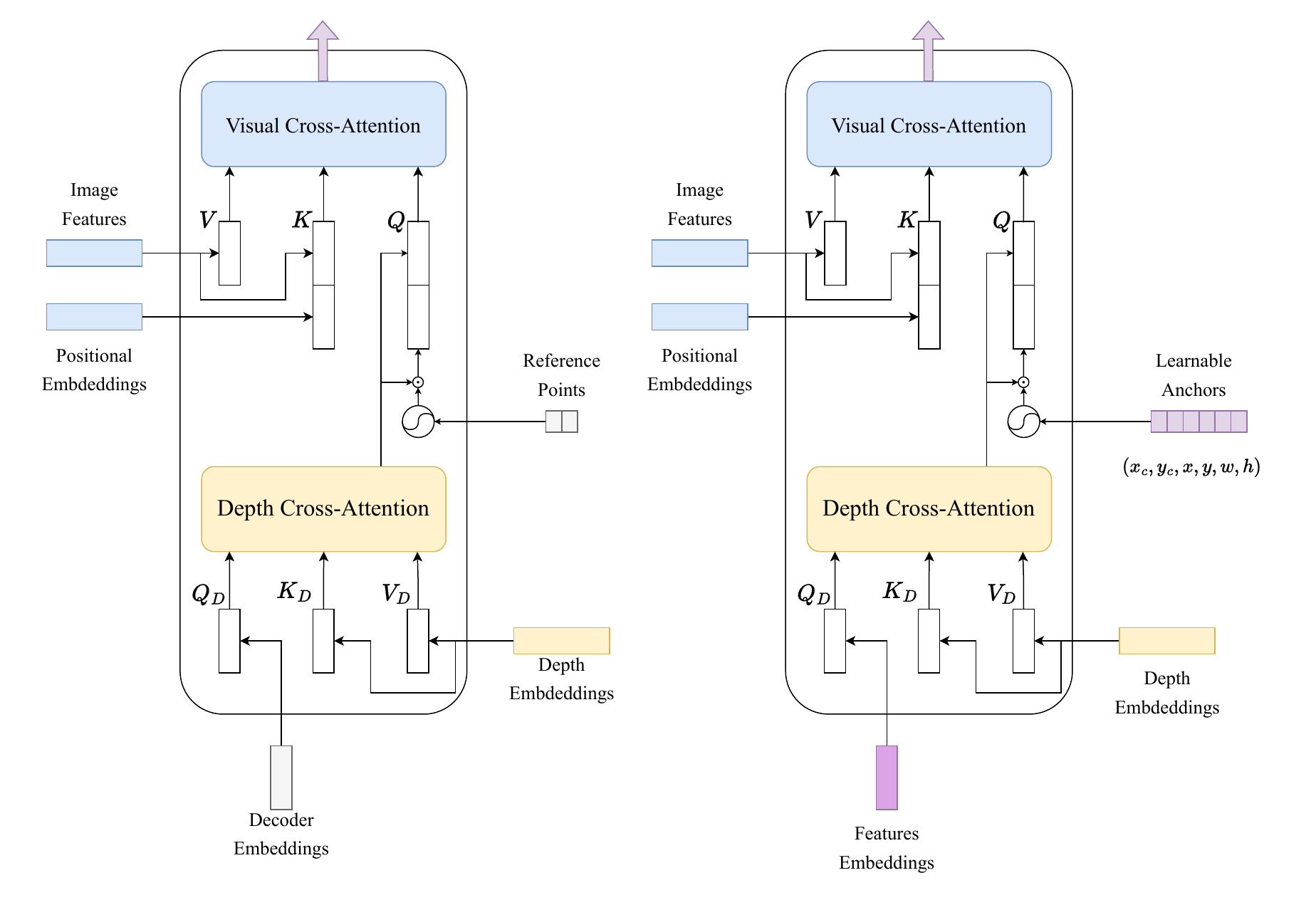}
	\caption{
		The differences in details between our decoder and MonoDETR's decoder, the main difference between the two being the source of the embedding and the presence or absence of an initial anchors.}
	\label{decoder}
\end{figure}

In Deformable-DETR \cite{zhu2020deformable}, object queries are a set of learnable embeddings optimized by the decoder and mapped to classification scores and bounding boxes via prediction heads. However, these object queries are challenging to interpret and optimize as they lack explicit physical meaning. Subsequent works \cite{li2022dn,wang2022anchor,liu2022dab,zhang2022dino,yao2021efficient} have improved the initialization of object queries and extended them to include content queries and anchors. In 3D object detection, the position of objects is no longer a simple 2D bounding box but includes depth information in a 3D bounding box. Therefore, a more accurate and interpretable method for generating 3D object queries is needed. To address this, we introduce the process of generating object queries from 2D anchors into monocular 3D object detection, expanding it into depth-aware object queries.


To put it simply, we have designed a target query generation method based on the characteristics of 3D objects. In MonoDETR\cite{zhang2023monodetr}, the final output of the bbox is a six-dimensional vector, $[x_c,y_c,x,y,w,h]$, $x_c,y_c$ represent the 3D center point of the object in the image coordinates, and $x,y,w,h$ define the 2D bbox. However, MonoDETR's object query generates only a two-dimensional vector, consisting of a set of learnable embeddings, an approach that is clearly not ideal. MV2D\cite{wang2023object} and QAF2D\cite{ji2024enhancing} provided us with inspiration, demonstrating that 2D bounding boxes can enhance the performance of 3D detection. Therefore, we incorporated the process of generating object queries from the 2D DETR model into our method, \(x, y, w, h\) are directly learned from the image features, while the 3D center coordinates \(x_c, y_c\) are obtained by fusing depth information \(f_D\) with visual features \(f_v\). Specifically, the following formulations are used:

\begin{align}
	\begin{aligned}
		&(x, y, w, h) = \text{MLP}_1(f_v) \\
		&(x_c, y_c) = \text{MLP}_2(\text{Attn}(f_v,f_D, f_D))
	\end{aligned}
\end{align} Here, \(\text{MLP}_1\) is used to predict the 2D bounding box parameters solely from the visual features, while \(\text{MLP}_2\) and Attn incorporate both the depth and visual features to estimate the 3D center coordinates. The inclusion of depth information in Attn allows the model to better capture 3D spatial relationships, leading to more accurate localization of the object's 3D center.

One more point is that the query embedding we use is also different from MonoDETR, our embedding comes directly from the image features, the specific difference is shown in Figure \ref{decoder}.

\subsubsection{Efficient Feature Extraction}
\label{encoder}


It is well-established that efficient feature extraction methods often enhance model performance. In DETR-type models, feature extraction is primarily carried out by the backbone and encoder components. We experimented with various advanced networks for both components.

Regarding the backbone, essentially all our attempts were unsuccessful. Neither ConvNeXt \cite{liu2022convnet}, Swin Transformer \cite{liu2021swin}, nor EfficientNet \cite{pham2021meta} outperformed ResNet \cite{he2016deep} in effectiveness. However, our efforts with the Encoder proved fruitful. After modifying RT-DETR's \cite{zhao2024detrs} Hybrid Encoder, we achieved performance improvements in our model. As illustrated in Figure \ref{framework}, the enhanced encoder in this study comprises two main components: a single encoder layer and a Cross-Scale Feature Integration Module (CFIM). The integration module's function is to combine adjacent features into a new feature representation, with its structure depicted in Figure \ref{submodel}. This fusion process is succinctly formalized by the ensuing equations:

\begin{align}
	\begin{gathered}
		\mathbf{Q}=\mathbf{K}=\mathbf{V}=\text { Flatten }\left(S_5\right) \\
		F_5=\operatorname{Reshape}(\operatorname{Attn}(\mathbf{Q}, \mathbf{K}, \mathbf{V})) \\
		\text { Output }=\operatorname{CFIM}\left(\left\{S_3, S_4, F_5\right\}\right)
	\end{gathered}
\end{align}

Compared to the hybrid encoder of RT-DETR, our hybrid encoder exhibits differences in the Fusion module.

\subsubsection{Overall Loss}

In order to make the anchors work better, we also calculate the loss using the anchors obtained in the \ref{query}, and the $\mathcal{L}_{enc}$ is as follows

\begin{align}
	\begin{aligned}
		\mathcal{L}_{enc}(\hat{y}, y) & =\mathcal{L}_{\text {box }}(\hat{b}, b)+\mathcal{L}_{c l s}(\hat{c}, \hat{b}, y, b) \\
		& =\mathcal{L}_{\text {box }}(\hat{b}, b)+\mathcal{L}_{cls}(\hat{c}, c, I o U)
	\end{aligned}
\end{align}
where $\hat{y}$ and $y$ denote prediction and ground truth, $\hat{y}={\hat{c}, \hat{b}}$ and $y={c, b}$, $c$  represent categories and $b$ represent bounding boxes and 3D centers, respectively.

Finally, our overall loss is similar to that of MonoDETR, except $\mathcal{L}_{enc}$:
\begin{align}
	\mathcal{L} _{\mathrm{overall}}=\frac{1}{N_{\mathrm{gt}}}\cdot \sum_{n=1}^{N_{gt}}{\left( \mathcal{L} _{2D}+\mathcal{L} _{3D}+\mathcal{L} _{enc} \right)}+\mathcal{L} _{\mathrm{dmap}}
\end{align}
where $\mathcal{L}_{d \text{map }}$ represents the focal loss\cite{lin2017focal} of the predicted categorical foreground depth map $D_{f g}$ in Section.\ref{depth_predictor} and $N_{gt}$ denotes the number of ground-truth objects. Additionally, it is noteworthy that the proportion of the total weight attributed to the $\mathcal{L}_{d \text{map}}$ loss during training differs between MonoDETRNext-A and MonoDETRNext-E, with MonoDETRNext-A having a higher percentage.

\subsection{MonoDETRNext-A}
The principal disparity between our proposed models, MonoDETRNext-E and MonoDETRNext-A, resides in their Depth Predictor architecture. The initial Depth Predictor in MonodDETR, by design, featured a lightweight configuration comprising solely two $3\time3$ convolutional layers, extracting $f_{D} \in \mathbb{R}^{\frac{H}{16} \times \frac{W}{16} \times(C)}$ from features preprocessed by the backbone. The accuracy of monocular 3D object detection correlates significantly with the quality of depth information obtained. In pursuit of refined depth estimation and subsequent enhancement of 3D detection precision, we undertook a redesign of the Depth Predictor. In order to further improve the accuracy of the model, we tried a variety of well-established depth estimation networks, and finally we chose the most effective one, the structure of this network we will show in this section.


\subsubsection{Accurate Depth Predictor}
\label{depth_predictor}

Similar to the feature extraction process discussed in the previous section, we experimented with multiple state-of-the-art monocular depth estimation models, including Lite-Mono\cite{zhang2023lite} , Depth-V2\cite{godard2019digging}, and ZoeDepth \cite{bhat2023zoedepth}. Ultimately, only a simplified version of Lite-Mono improved detection performance. Conversely, other models led to a decline in detection accuracy. We posit that this outcome may be attributed to the challenge of balancing network training. It is crucial to note that our primary task is detection, not depth estimation. Excessively large depth estimation networks may potentially disrupt the training process, consequently leading to diminished detection performance.

The architectural of this Depth Predictor draws is depicted in Figure \ref{framework}, initiates with the processing of images of dimensions $H \times W \times 3$ through a convolutional perturbation module, where image data undergoes downsampling via a $3 \times 3$ convolutional operation. Subsequently, two additional $3 \times 3$ convolutional layers, each with a stride of $1$, are deployed for local feature extraction, yielding feature maps of dimensions $\frac{H}{2} \times \frac{W}{2} \times C_1$. In the ensuing stage, these features are concatenated with the pooled three-channel input image, followed by another downsampling step through a $3 \times 3$ convolutional layer with a stride of $2$, resulting in feature maps of dimensions $\frac{H}{4} \times \frac{W}{4} \times C_2$. , RGFI and SDC are introduced to facilitate the acquisition of rich hierarchical feature representations. The ensuing downsampling layers also inherit connected features from the preceding downsampling layer. Analogously, the output feature maps undergo further processing until their dimensions align with those of MonoDETR\cite{zhang2023monodetr}, specifically $\frac{H}{16} \times \frac{W}{16} \times C_4$.
\begin{figure}
	\centering
	\includegraphics[width=1\columnwidth]{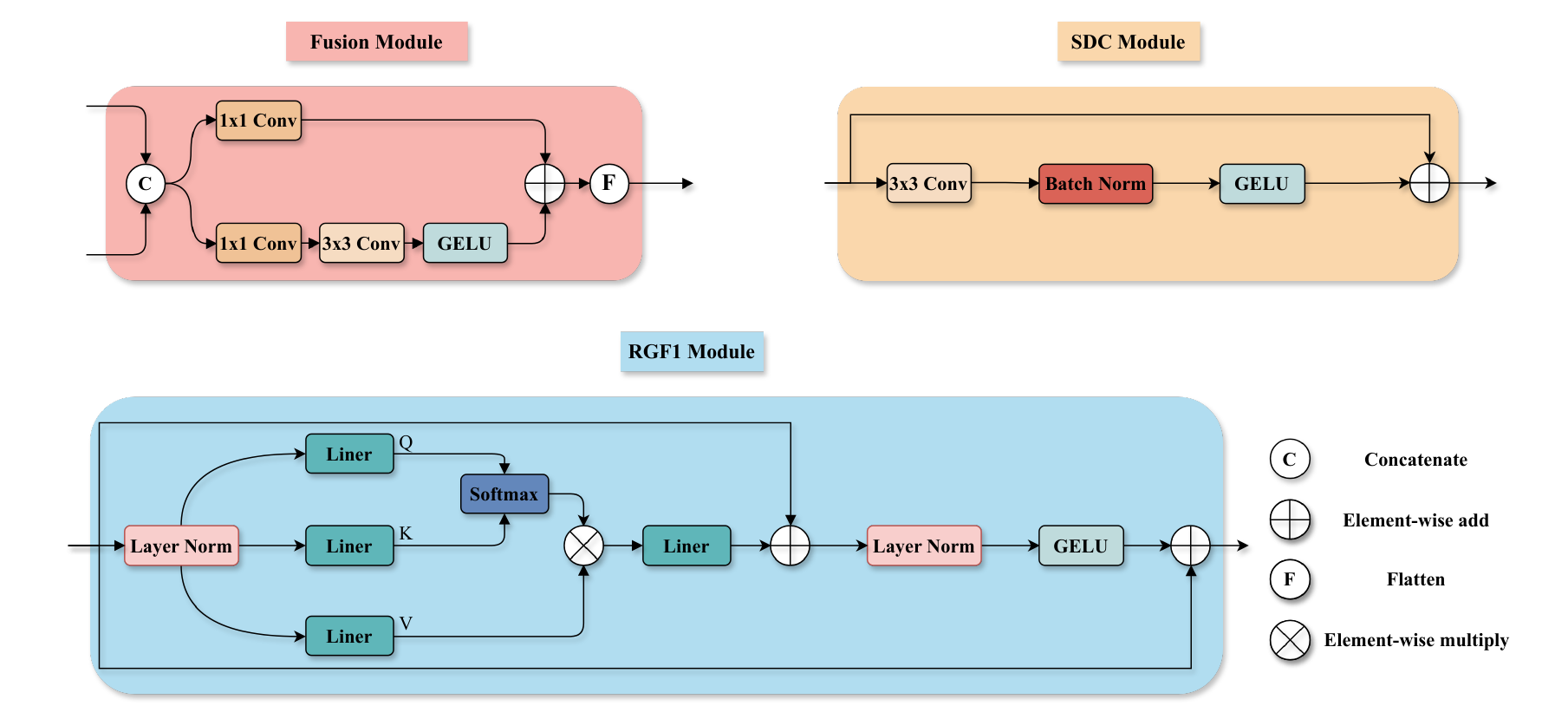}
	\caption{
		The fusion block within CFIM is depicted, showcasing the architectures of the Sequential Dilated Convolution (SDC) module and the Regional-Global Feature Interaction (RGFI) module proposed therein.}
	\label{submodel}
\end{figure}


\textbf{The Sequential Dilated Convolution (SDC)} module is introduced to extract multi-scale local features using dilated convolutions. Similar to lite-mono, we adopt a staged approach by inserting multiple consecutive dilated convolutions with varying dilation rates to effectively aggregate multi-scale context. Compared to the CDC module of Lite-Mono, our SDC module has not point-wise layers, resulting in a more lightweight architecture.

Given a 2D signal $x[i]$, the output $y[i]$ of a 2D dilated convolution can be defined as:
\begin{align}
	y[i]=\sum_{k=1}^K x[i+r \cdot k] w[k],
\end{align}
where $w[k]$ is a filter of length $K$, and $r$ denotes the dilation rate used for convolving the input $x[i]$. In standard non-dilated convolutions, $r=1$. By employing dilated convolutions, the network can maintain a fixed size for the output feature map while achieving a larger receptive field. Considering an input feature $X$ with dimensions $H \times W \times C$, the output $\hat{X}$ of our SDC module is computed as follows:

\begin{align}
	\hat{X}=X+G \left(\left(B N\left( \text { Conv }_r(X)\right)\right)\right),
\end{align}

where $G$ denotes the GELU\cite{hendrycks2016gaussian} activation. $BN$ represents the batch normalization layer, and $\text{Conv}_r(\cdot)$ is a $3 \times 3$ convolutional layer.

\textbf{The Regional-Global Feature Interaction (RGFI)} module operates as follows: given an input feature map $X$ of dimensions $H \times W \times C$, it is linearly projected into queries $Q=X W_q$, keys $K=X W_k$, and values $V=X W_v$, where $W_q$, $W_k$, and $W_v$ are weight matrices. Cross-covariance attention\cite{ali2021xcit} is employed to enhance the input $X$:

\begin{align}
	\widetilde{X}=\operatorname{Attention}(Q, K, V)+X,
\end{align}

Subsequently, non-linearity is introduced to enhance the features:
\begin{align}
	\hat{X}=X+ G (\operatorname{LN}(\tilde{X})),
\end{align}

where $LN$ represents a layer normalization\cite{ba2016layer} operation. 
Similar to the SDC module, our RGF1 module does not include point-wise layers, unlike the LGF1 module of Lite-Mono \cite{zhang2023lite}. The complete process is illustrated in Figure \ref{submodel} .

\section{Experiments}

\subsection{Setting}
\subsubsection{Dataset}
\begin{table*}[htbp]
	\centering 
	\renewcommand\arraystretch{1.1}
	\caption{\textbf{Monocular performance of the car category on KITTI test and val sets.} We utilize bold numbers to highlight the best results, and color the second-best ones expect our methods and our gain over them in blue. The FPS values are obtained through speed testing conducted on individual models using a single NVIDIA GeForce RTX 3090 GPU. During testing, a batch size of 1 was utilized.}
	\label{kitti_comparison}
	\begin{tabular}{l|c|ccc|ccc|ccc}
		\hline
		\multirow{2}{*}{Method} & \multirow{2}{*}{$FPS_{bz=1}$} & \multicolumn{3}{c|}{Test, $AP_{3D}$} & \multicolumn{3}{c|}{Test, $AP_{BEV}$} & \multicolumn{3}{c}{Val, $AP_{3D}$} \\
		& & Easy & Mod. & Hard & Easy & Mod. & Hard & Easy & Mod. & Hard \\
		\midrule
		
		SMOK\cite{liu2020smoke} & 17.6 & 14.03 & 9.76 & 7.84 & 20.83 & 14.49 & 12.75 & 14.76 & 12.85 & 11.50 \\
		MonoPair \cite{Chen_2020_CVPR} & - & 13.04 & 9.99 & 8.65 & 19.28 & 14.83 & 12.89 & 16.28 & 12.30 & 10.42 \\
		RTM3D \cite{li2020rtm3d} & 23.3 & 13.61 & 10.09 & 8.18 & - & - & - & 19.47 & 16.29 & 15.57 \\
		PatchNet \cite{ma2020rethinking} & - & 15.68 & 11.12 & 10.17 & 22.97 & 16.86 & 14.97 & - & - & - \\
		D4LCN \cite{ding2020learning} & - & 16.65 & 11.72 & 9.51 & 22.51 & 16.02 & 12.55 & - & - & - \\
		DDMP-3D \cite{wang2021depth} & 9.4 & 19.71 & 12.78 & 9.80 & 28.08 & 17.89 & 13.44 & - & - & - \\
		MonoRUn \cite{chen2021monorun} & 20.2 & 19.65 & 12.30 & 10.58 & 27.94 & 17.34 & 15.24 & 20.02 & 14.65 & 12.61 \\
		CaDDN \cite{reading2021categorical} & - & 19.17 & 13.41 & 11.46 & 27.94 & 18.91 & 17.19 & 23.57 & 16.31 & 13.84 \\
		
		PGD \cite{wang2022probabilistic} & 23.9 & 19.05 & 11.76 & 9.39 & 26.89 & 16.51 & 13.49 & 19.27 & 13.23 & 10.65 \\
		MonoDLE \cite{ma2021delving} & 25.0 & 17.23 & 12.26 & 10.29 & 24.79 & 18.89 & 16.00 & 17.45 & 13.66 & 11.68 \\
		MonoRCNN \cite{Shi_2021_ICCV} & 24.7 & 18.36 & 12.65 & 10.03 & 25.48 & 18.11 & 14.10 & 16.61 & 13.19 & 10.65 \\
		MonoGeo \cite{9665911} & - & 18.85 & 13.81 & 11.52 & 25.86 & 18.99 & 16.19 & 18.45 & 14.48 & 12.87 \\
		GUPNet \cite{lu2021geometry} & 28.7 & 20.11 & 14.20 & 11.77 & - & - & - & 22.76 & 16.46 & 13.72 \\
		
		MonoDTR \cite{huang2022monodtr} & 27.0 & 21.99 & 15.39 & 12.73 & 28.59 & 20.38 & 17.14 & 24.52 & 18.57 & 15.51 \\
		MonoUNI \cite{jinrang2024monouni} & - & 24.75& 16.73& 13.49 & - & - & - & 24.66 & 17.18 & 14.06 \\
		AutoShape \cite{liu2021autoshape} & - & 22.47 & 14.17 & 11.36 & 30.66 & 20.08 & 15.59 & 20.09 & 14.65 & 12.07 \\
		MonoDETR \cite{zhang2023monodetr} & 26.5 & 25.00& 16.47 & 13.58 & 33.60 & 22.11 & 18.60 & \textcolor{blue}{28.84} & \textcolor{blue}{20.61}& \textcolor{blue}{16.38} \\
		
		MonoCD \cite{yan2024monocd} & \textcolor{red}{31.7} & 25.53& 16.59& 14.53 &33.41& 22.81& 19.57 & 26.45& 19.37& 16.38\\
		DA3D \cite{10497146} & - & \textcolor{blue}{27.76}& \textcolor{blue}{20.47}& \textcolor{blue}{17.90} &\textcolor{red}{36.83} &\textcolor{blue}{26.92}& \textcolor{blue}{23.41} & - & - &- \\
		 \midrule
		MonoDETRNext-E & \textcolor{blue}{29.2} & 27.35 &	21.24 &	19.16  & 34.73 & 27.32 &	24.87   & 30.28& 24.77& 20.38\\
		
		MonoDETRNext-A & 20.8 &	\textcolor{red}{28.52} &	\textcolor{red}{22.81} &	\textcolor{red}{21.08}  & \textcolor{blue}{35.86} & \textcolor{red}{28.96} &\textcolor{red}{27.58} & \textcolor{red}{32.95}& \textcolor{red}{25.01}& \textcolor{red}{21.92} \\
		\hline
		
	\end{tabular}
\end{table*}

\textbf{KITTI}\cite{geigerAreWeReady2012} encompasses 7,481 training and 7,518 test images. Following the protocol outlined in previous studies \cite{chen2016monocular,chen20153d}, we segregate 3,769 validation images from the training set. Evaluation is conducted across three difficulty levels: easy, moderate, and hard. Performance metrics are computed based on average precision ($AP$) of bounding boxes in both 3D space ($AP_{3D}$) and the bird's-eye view ($AP_{BEV}$). These metrics are assessed at 40 recall positions to ensure comprehensive evaluation.

\textbf{Rope3D}\cite{ye2022rope3d} benchmark is designed for autonomous driving and monocular 3D object detection. It consists of approximately 50,000 images and contains over 1.5 million 3D objects. These images capture various real-world environments, with diverse camera settings and environmental conditions, making it a highly diverse and challenging dataset for roadside perception tasks. For the evaluation metrics, we follow the official settings using $AP_{3D}$ and $Rope_{score}$ \cite{ye2022rope3d}, which is a combined metric of the 3D AP and other similarities, such as Average Ground Center Similarity (ACS). We follow the proposed homologous setting to utilize 70$\%$ of the images as training, and the remaining as validating.

\subsubsection{End-to-End 3D object tracking}
To further demonstrate versatility, we also extend MonoDETRNext to 3D Multi-Object Tracking (MOT), the specific tracking approach is consistent with MOTR \cite{zeng2022motr}, interested readers can read it by themselves, here we use the same training method as \cite{pan2023mo}. Through this experiment, we demonstrate that the query-based MOT model has higher accuracy in 3d tracking when using the 3d detection model as the base model. The primary metrics employed for evaluation included HOTA, AssA and MOTA \cite{ristani2016performance,luiten2021hota}. These metrics collectively provided a comprehensive and robust evaluation of our model's tracking performance.

\subsubsection{Implementation details}
We employed ResNet50\cite{he2016deep} as our backbone. We utilized eight attention heads for all attention modules, including RGFI. MonoDETRNext-E was trained for 230 epochs on a single RTX 4090 GPU, with a batch size of 8 and a learning rate of $2 \times 10^{-4}$. MonoDETRNext-A was trained for 300 epochs. We employed the AdamW \cite{loshchilov2017decoupled} optimizer with weight decay of $10^{-4}$. For MonoDETRNext-E, the learning rate was decreased by a factor of 0.1 at epochs 145 and 205 and 
for MonoDETRNext-A, the process is between 165 and 205. For the Rope3D, with 150 epochs of both the MonoDETRNext-E and A trained, the learning rate decreases starting at 80 epochs and ending at 120 epochs. Tracking, on the other hand, was trained for another 200 epochs on the MOT part of the KITTI dataset using the E version on top of the detection The number of object queries was consistent with MonoDETR, set at 50. Inference speed was evaluated on a single RTX 3090 GPU.

\subsection{Main Results}

\textbf{KITTI:} The table \ref{kitti_comparison} presents a comparison of various monocular 3D object detection models on the KITTI dataset, focusing on the car category's performance. The evaluation metrics  $AP_{3D}$ and $AP_{BEV}$ , which are both at 40 recall positions, across different difficulty levels (Easy, Moderate, and Hard) for both the test and validation sets. Additionally, the FPS metric for batch size 1 is provided to gauge the inference speed of each model. 

Among the methods listed, our methods MonoDETRNext-E and MonoDETRNext-A stands out. These variants exhibit notable advantages over other models in terms of both accuracy and efficiency. In terms of accuracy, MonoDETRNext-A achieves impressive $AP_{3D}$ and $AP_{BEV}$ scores across all difficulty levels, surpassing all other models listed. Specifically, it outperforms MonoDETR in $AP_{3D}$, showcasing its superior ability to accurately detect 3D objects. Additionally, it achieves the highest $AP_{BEV}$ scores, indicating its effectiveness in detecting objects from a bird's eye view perspective. At the same time, we found that our method significantly outperforms other methods in both moderate and hard scenarios, which is probably because the introduction of anchors significantly improves the impact of occlusion on detection performance. Furthermore, \text{MonoDETRNext-E} exhibits respectable inference speeds, with \text{MonoDETRNext-E} achieving the second $FPS_{bz=1}$ among all models.


\textbf{Rope3D:} For the Rope3D performance comparison in Table \ref{result_rope3d}, MonoDETRNext achieves \textit{state of-the-art} results, with the MonoDETRNext-A variant surpassing all other methods, including BEVDepth and MonoFlex, in both 3D object detection ($AP_{3D}$ of 82.34) and $Rope_{score}$ (85.16). This highlights the model's superiority in 3D scene understanding from monocular images.

\begin{table}
	\begin{center}
		\caption{Performance comparison between MonoDETRNext and existing methods on the Rope3D \cite{ye2022rope3d} set.}
		\label{result_rope3d}
			
			\begin{tabular}{l|cc}
				\hline Method & $AP_{3D}$  & $Rope_{score}$         \\
				\noalign{\smallskip}
				\hline
				\noalign{\smallskip}
				MonoFlex\cite{zhang2021objects}  &  59.78 &66.66   \\
				GUPNet\cite{lu2021geometry}  &  66.52& 70.14 \\
				BEVDepth\cite{li2023bevdepth}  & 69.63& 74.70    \\
				
				BEVHeight\cite{yang2023bevheight}  & 74.60& 78.72    \\
				MonoDLE \cite{ma2021delving}  & 77.50 & 80.84    \\
				MonoDETR\cite{zhang2023monodetr} & 78.68 & 81.93  \\
				\hline
				MonoDETRNext-E & 81.89 & 83.69  \\
				MonoDETRNext-A & \textbf{82.34} &  \textbf{85.16} \\
				\noalign{\smallskip}
				\hline
			\end{tabular}
	\end{center}
\end{table}

\textbf{Tracking:} As shown in Table \ref{result_Tracking}, the tracking results indicate that MonoDETRNext-E-MOT sets a new benchmark on the KITTI, outperforming other end-to-end and non-end-to-end methods like CenterTr and PolarMOT. It achieves high HOTA, MOTA, and AssA, as well as relatively low IDSW, indicating its robustness and accuracy in target tracking.  This suggests MonoDETRNext’s strong potential in both object detection and tracking tasks, offering a unified and efficient solution across multiple benchmarks.

\begin{table}
	\begin{center}
		\caption{The tracking performance on KITTI after converting MonoDETRNext into an end-to-end tracking model.}
		\label{result_Tracking}
		\setlength{\tabcolsep}{2pt}
		\resizebox{\columnwidth}{!}{
			
			\begin{tabular}{l|ccccc}
				\hline Tracker & HOTA $\uparrow$ & MOTA$\uparrow$  & AssA$\uparrow$ &DetA$\uparrow$ & IDSW$\downarrow$   \\
				\noalign{\smallskip}
				\hline
				\noalign{\smallskip}
				\textit{Non-End-to-end:} &&&&&\\
				CenterTr \cite{zhou2020tracking}  & 73.02 & 88.83 & 71.12 &75.62&254 \\
				PolarMOT \cite{kim2022polarmot}  & 	75.16 & 85.08  & 76.95&	73.94&462 \\
				TripletTrack \cite{marinello2022triplettrack}  & 	73.58 & 	84.32 & 74.66&73.18& 322 \\
				PermaTr \cite{tokmakov2021learning}  & 	78.03 & 90.13  & 	78.41&	\textbf{78.29} &258 \\
				
				\noalign{\smallskip}
				\hline
				\noalign{\smallskip}
				\textit{End-to-end:} &&&&&\\
				DecoderTracker\cite{pan2023mo} & 72.08 & 83.19 & 73.84 &71.84&275 \\
				MonoDETRNext-E-MOT  & \textbf{78.93} & \textbf{90.27} & \textbf{81.11} &73.05& \textbf{143} \\
				
				\noalign{\smallskip}
				\hline
			\end{tabular}
		}
	\end{center}
\end{table}

\subsection{Ablation Studies}
\label{ablation_studies}

We have validated the effectiveness of each component and strategy, and reported the $AP_{3D}$ for the car category on the KITTI validation benchmark.


Our study evaluates different strategies for object query generation, as depicted in Table \ref{ablation_results_query}. "L-center assign learnable embeddings," "enc-box," and "enc-center" refer to the anchors derived from backbone features, generated separately by two heads and contributing to individual loss calculation. From the table, it is evident that MonoDETRNext significantly improves detection performance in moderate and challenging conditions. This improvement suggests that the strategy of using queries to generate anchors can mitigate some of the adverse effects caused by object occlusions.
\begin{table}
	\centering
	\centering
	\captionsetup{justification=raggedright, singlelinecheck=false} 
	\caption{Performance of MonoDETRNext-E at different 3D object query generation strategy.}
	\label{ablation_results_query}
	\setlength{\tabcolsep}{6pt}
	\begin{tabular}{l|ccc}
		\toprule
		Stage & Easy & Mod. & Hard \\
		\midrule
		L-center               & 25.80& 18.96& 17.47 \\
		L-center+enc-box         & 27.89& 21.49& 19.96 \\
		enc-center                 & 28.89& 20.78& 18.52 \\
		enc-center+enc-box       & 30.28& 24.77& 20.38  \\
		\bottomrule
	\end{tabular}
\end{table}

Table \ref{ablation_results_encoder} illustrates the performance of MonoDETRNext-E with different vision encoders, including the "hybrid" encoder featured in Figure \ref{framework}. Notably, the "hybrid" encoder demonstrates superior performance across Easy, Moderate, and Hard categories compared to other configurations. Interestingly, even with fewer layers, the "hybrid-L" encoder maintains competitive performance, suggesting that a lower layer count may be more effective. 
\begin{table}
	\centering
	\captionsetup{justification=raggedright, singlelinecheck=false} 
	\caption{Performance of MonoDETRNext-E under different vision encoder.}
	\label{ablation_results_encoder}
	\setlength{\tabcolsep}{6pt}
	\begin{tabular}{l|ccc}
		\toprule
		Model & Easy & Mod. & Hard \\
		\midrule
		RT-DETR's   & 26.42& 20.45& 16.78 \\ 
		MonoDETR's   & 28.35& 22.53& 18.74 \\ 
		hybrid-L   & 29.67& 23.81& 19.03 \\ 
		our hybrid    & 30.28& 24.77& 20.38\\ 
		\bottomrule
	\end{tabular}
\end{table}

In Table \ref{ablation_results_depth}, $N_1$, $N_2$, and $N_3$ signify the number of SDC modules, as depicted in Figure \ref{framework}. The configuration $(2, 2, 4)$ yields the best results across all levels, indicating an optimal balance. Both smaller $(1, 1, 2)$ and larger scales $(3, 3, 6)$ and $(4, 4, 6)$ show diminished performance, suggesting issues with underfitting and overfitting respectively.
\begin{table}
	\centering
	\centering
	\caption{Effect of Different Scales of Accurate Depth Predictors on MonoDETRNext-A Performance.}
	\label{ablation_results_depth}
	\setlength{\tabcolsep}{6pt}
	\begin{tabular}{c|ccc}
		\toprule
		$N_1,N_2,N_3$ & Easy & Moderate & Hard \\
		\midrule
		2,   2, 4 & 32.95& 25.01& 21.92 \\
		1,   1, 2 & 30.26& 24.99& 22.58 \\
		3,   3 ,6 & 32.45& 23.85& 21.85 \\
		4,   4, 6 & 31.78& 23.92& 20.78 \\
		\bottomrule
	\end{tabular}
\end{table}

The Table.\ref{ablation_results_depth_module} presents an ablation study on the depth estimation network module in MonoDETRNext-A. Removing the point-wise operations improves performance across all difficulty levels, particularly evident in the "Easy" category, highlighting their significant contribution to depth estimation accuracy.

\begin{table}
	\centering
	\centering
	\caption{The impact of removing point-wise in the depth estimation network module on the performance of MonoDETRNext-A. \Checkmark indicates deletion, and \XSolidBrush indicates no deletion.}
	\label{ablation_results_depth_module}
	\setlength{\tabcolsep}{6pt}
	\begin{tabular}{cc|ccc}
		\toprule
		SDC &RGFI& Easy & Moderate & Hard \\
		\midrule
		\XSolidBrush& \XSolidBrush& 29.62& 22.43& 18.86 \\
		$\textbf{\Checkmark}$& \textbf{\XSolidBrush}& 31.86& 22.86& 22.04 \\
		\XSolidBrush& $\textbf{\Checkmark}$& 31.54& 23.42& 21.81 \\
		$\textbf{\Checkmark}$& $\textbf{\Checkmark}$& 32.95& 25.01& 21.92 \\
		\bottomrule
	\end{tabular}
\end{table}

In Table \ref{ablation_results_position}, the impact of different depth positional encodings on MonoDETRNext-E is examined. "3D sin/cos" and "2D sin/cos" utilize sinusoidal functions for encoding, with the former incorporating all six output dimensions. Notably, "3D sin/cos" outperforms other settings across Easy, Moderate, and Hard categories. This suggests that the precise encoding of depth information through sinusoidal functions significantly benefits object detection. Conversely, "Meter-wise" and "k-bin" positional encodings yield comparatively lower performance, indicating less effective depth representation in these configurations. 
\begin{table}
	\centering
	\captionsetup{justification=raggedright, singlelinecheck=false} 
	\caption{The impact of different depth positional embeddings of MonoDETRNext-E. }
	\label{ablation_results_position}
	\setlength{\tabcolsep}{4pt}
	\begin{tabular}{l|ccc}
		\toprule
		Settings & Easy & Mod. & Hard \\
		\midrule
		3D sin/cos   & 30.28& 24.77& 20.38\\
		2D sin/cos   & 28.21& 21.52& 18.77\\
		Meter-wise   & 26.29& 20.04& 18.13\\
		k-bin & 25.20& 18.51& 16.85\\
		
		\bottomrule
	\end{tabular}
\end{table}

\section{Conclusion and Limitations}
\textbf{Conclusion}: This paper introduces a novel monocular vision-based approach for 3D object detection. Leveraging advancements from the 2D detection domain, we propose the efficient and precise MonoDETRNext. Building upon the groundwork laid by MonoDETR, we introduce two variants: MonoDETRNext-E prioritizing speed, and MonoDETRNext-A emphasizing accuracy. Our methodology encompasses the development of an efficient hybrid vision encoder, enhancements to depth prediction mechanisms, and refinements in object query generation. Through comprehensive performance evaluation, we establish the superiority of our model over existing methods. By optimizing both accuracy and computational efficiency, MonoDETRNext sets a new benchmark in monocular 3D object detection, facilitating future research and applications across diverse real-world scenarios.

\textbf{Limitations}: Despite the substantial advancements achieved by MonoDETRNext in enhancing the accuracy and computational efficiency of monocular 3D object detection, certain limitations persist. Due to the inherent constraints of monocular vision methods, there remains a notable discrepancy in accuracy and performance compared to approaches employing multi-view methodologies or sensor fusion techniques, such as the integration of LiDAR with cameras.
{
    \small
    \bibliographystyle{ieeenat_fullname}
    \bibliography{natbib}
}

\clearpage
\maketitlesupplementary

\begin{figure*}
	\centering
	\includegraphics[width=1\textwidth]{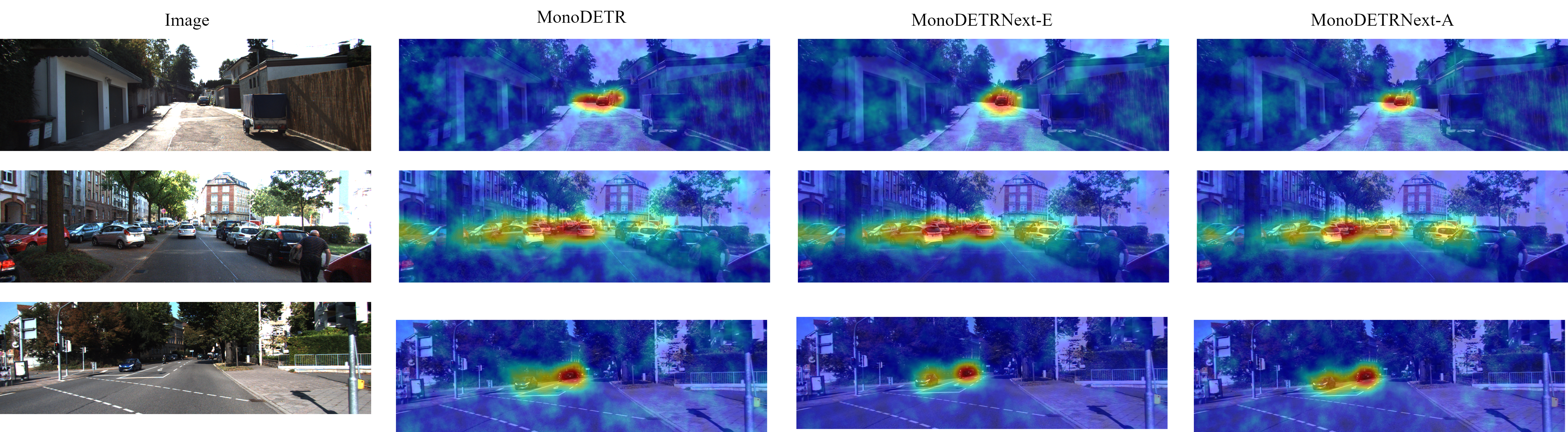}
	\caption{
		The visualization results of the final attention output from the decoder of MonoDETR and MonoDETRNext. Hotter colors indicate higher attention weights. It is evident that the heat map of MonoDETRNext is more concentrated than that of MonoDETR.}
	\label{hotmap}
\end{figure*}
\section{Some discussions on the success of MonoDETRNext}

The success of MonoDETRNext can be attributed to a comprehensive understanding of the limitations inherent in existing models, particularly MonoDETR, and the innovative strategies employed to address these challenges. By critically analyzing the shortcomings of MonoDETR, the authors identified key areas for improvement, leading to significant advancements in monocular 3D object detection.

One of the most notable enhancements in MonoDETRNext is the optimization of query initialization. In the original MonoDETR framework, query initialization posed constraints that adversely affected detection efficiency and accuracy. MonoDETRNext tackles this issue by implementing a refined initialization strategy that allows the model to more effectively capture crucial scene features, thus enhancing overall detection performance. This improvement demonstrates the importance of fine-tuning initial conditions to bolster model efficacy in dynamic environments.

The introduction of a more powerful depth estimator is another cornerstone of MonoDETRNext’s advancements. This enhanced estimator significantly increases accuracy in depth perception, allowing the model to construct a more precise understanding of three-dimensional structures within a scene. By leveraging advanced deep learning techniques, the new depth estimator minimizes errors in complex environments, contributing to higher overall detection quality.

In Figure \ref{hotmap}, we present the visualization heatmaps of the attention outputs from our two models and MonoDETR. A closer examination reveals that our models exhibit a more concentrated and accurate attention distribution in key areas. This indicates that our models are more effective in identifying and focusing on critical information within the images, thereby enhancing detection accuracy.

In conclusion, the achievements of MonoDETRNext underscore the significance of addressing fundamental issues within existing frameworks. Through innovative enhancements in query initialization, encoder architecture, and depth estimation, the model sets a new standard for monocular 3D object detection. The accompanying heatmap analyses provide compelling evidence of these advancements, affirming the model's potential to influence future research and applications in the field.

\section{How to achieve better detection accuracy on KITTI}

It is well known that one of the best ways to enhance the performance of deep learning models is by increasing the dataset size. In pursuit of maximizing the performance of MonoDETRNext, we attempted to expand the dataset.

The original Repo3D dataset was a reasonable choice; however, its primary focus on traffic signals does not align well with the scenarios found in the KITTI dataset. Therefore, we turned our attention to another dataset commonly used in autonomous driving, nuScenes. The nuScenes dataset contains a wealth of driving scenarios, encompassing urban streets, rural roads, and various weather conditions, making it highly suitable for training and evaluating 3D object detection models related to autonomous driving.

The nuScenes dataset comprises data from six cameras, and we selected the forward camera as our supplementary data source. Additionally, since the nuScenes dataset does not allow for the calculation of the $\alpha$ angle when converting to KITTI format, we only utilized it to supervise the depth module. One more thing ,similar to LPCG\cite{peng2022lidar}, we generate pseudo labels using KITTI raw data (42K unlabeled data) to add additional training data, further improving the performance of our method. Consequently, we focused our efforts on MonoDETRNext-A. The results are presented in Table \ref{kitti_best}.

\begin{table*}[htbp]
	\centering 
	\renewcommand\arraystretch{1.1}
	\caption{\textbf{Monocular performance of the car category on KITTI test and val sets.} The top three results are highlighted in the order of \textcolor{red}{red}, \textcolor{green}{green}, and \textcolor{blue}{blue}.}
	\label{kitti_best}
	\begin{tabular}{l|ccc|ccc}
		\hline
		\multirow{2}{*}{Method} & \multicolumn{3}{c|}{Test, $AP_{3D}$} & \multicolumn{3}{c}{Test, $AP_{BEV}$} \\
		& Easy & Mod. & Hard & Easy & Mod. & Hard \\
		\midrule
		MonoPair \cite{Chen_2020_CVPR} & 13.04 & 9.99 & 8.65 & 19.28 & 14.83 & 12.89 \\
		
		DDMP-3D \cite{wang2021depth} & 19.71 & 12.78 & 9.80 & 28.08 & 17.89 & 13.44 \\
		
		MonoRUn \cite{chen2021monorun} & 19.65 & 12.30 & 10.58 & 27.94 & 17.34 & 15.24 \\
		
		CaDDN \cite{reading2021categorical} & 19.17 & 13.41 & 11.46 & 27.94 & 18.91 & 17.19 \\
		
		MonoDLE \cite{ma2021delving} & 17.23 & 12.26 & 10.29 & 24.79 & 18.89 & 16.00 \\
		
		MonoRCNN \cite{Shi_2021_ICCV} & 18.36 & 12.65 & 10.03 & 25.48 & 18.11 & 14.10 \\
		
		MonoGeo \cite{9665911} & 18.85 & 13.81 & 11.52 & 25.86 & 18.99 & 16.19 \\
		
		GUPNet \cite{lu2021geometry} & 20.11 & 14.20 & 11.77 & - & - & - \\
		
		MonoDTR \cite{huang2022monodtr} & 21.99 & 15.39 & 12.73 & 28.59 & 20.38 & 17.14 \\
		
		AutoShape \cite{liu2021autoshape} & 22.47 & 14.17 & 11.36 & 30.66 & 20.08 & 15.59 \\
		
		MonoDETR\cite{zhang2023monodetr} & 25.00 & 16.47 & 13.58 & 33.60 & 22.11 & 18.60 \\
		
		SSD-MonoDETR\cite{he2023ssd} & 24.52 & 17.88 & 15.73 & 33.59 & 24.35 & 21.98 \\

		MonoTAKDv2 \cite{liu2024monotakd} & 29.86 &	21.26 &	18.27 & \textcolor{red}{43.83} &	\textcolor{red}{32.31} &	\textcolor{red}{28.48}  \\
		DA3D* \cite{10497146} & 	\textcolor{red}{30.83} &	\textcolor{blue}{22.08} &	\textcolor{blue}{19.20} &39.50 &	28.71 &	25.20  \\
		\midrule
		MonoDETRNext-E &27.35 &	21.24 &	19.16  & 34.73 & 27.32 &	24.87 \\
		
		MonoDETRNext-A & \textcolor{green}{28.52} & \textcolor{green}{22.81} & \textcolor{green}{21.08} & \textcolor{green}{35.86} & \textcolor{green}{28.96} & \textcolor{green}{27.58} \\
		
		MonoDETRNext-A(Extra Data) & \textcolor{blue}{29.94} & \textcolor{red}{24.14} & \textcolor{red}{23.79} & \textcolor{blue}{37.32} & \textcolor{blue}{30.68} & \textcolor{blue}{30.29} \\
		\hline
	\end{tabular}
\end{table*}

\end{document}